\documentclass[10pt, logo, copyright]{nv}

\usepackage{booktabs} 
\usepackage{caption}  
\usepackage{pifont}

\usepackage{amssymb}
\usepackage{subcaption}
\usepackage[table]{xcolor}
\usepackage{xcolor}   
\usepackage{multirow}
\usepackage{makecell} 
\usepackage{url}
\usepackage{booktabs}
\usepackage{graphicx}
\usepackage{animate}
\usepackage{xspace}
\usepackage{amsmath,amsfonts,bm}
\usepackage{enumitem}
\usepackage{cite}
\usepackage[ruled]{algorithm2e}

\usepackage{cleveref}
\crefname{section}{Sec.}{Secs.}
\crefname{table}{Tab.}{Tabs.}
\crefname{figure}{Fig.}{Figs.}
\crefname{algorithm}{Alg.}{Algs.}

\definecolor{cvprblue}{rgb}{0.21,0.49,0.74}
\definecolor{iccvblue}{rgb}{0.21,0.49,0.74}
\definecolor{mitblue}{rgb}{0.88,0.95,0.96}
\definecolor{gold}{rgb}{0.75,0.6,0.12}
\colorlet{shadecolor}{gray!40}
\definecolor{mydarkred}{rgb}{0.8,0.02,0.02}

\def\tablecite#1{[\citenum{#1}]}
\newcolumntype{g}{>{\columncolor{mitblue}}c}
\newcolumntype{f}{>{\columncolor{mitblue}}l}
\newcolumntype{h}{>{\columncolor{mitblue}}r}
\newcolumntype{i}{>{\columncolor{gray}}c}

\newcommand{\cmark}{\ding{51}}%
\newcommand{\xmark}{\ding{55}}%

\newcommand{\eg}{\textit{e.g.,}\xspace}
\newcommand{\ie}{\textit{i.e.,}\xspace}
\newcommand{\myPara}[1]{\noindent\textbf{#1}}

\usepackage[most]{tcolorbox}

\title{
  AnyFlow: Any-Step Video Diffusion Model with On-Policy Flow Map Distillation
}

\author{
  {\small Yuchao Gu$^{1,2}$, Guian Fang$^{2}$, Yuxin Jiang$^{2}$, Weijia Mao$^{2}$, Song Han$^{1,3}$,} \\ {\small \textbf{Han Cai$^{1*}$, Mike Zheng Shou$^{2*}$}} \\~\\
  $^{1}$NVIDIA \\
  $^{2}$Show Lab, National University of Singapore \\
  $^{3}$MIT\\
  \url{https://nvlabs.github.io/AnyFlow}
}
\correspondingauthor{Han Cai (hcai@nvidia.com), Mike Zheng Shou (mike.zheng.shou@gmail.com)}

\begin{abstract}
  Few-step video generation has been significantly advanced by consistency distillation.
  However, the performance of consistency-distilled models often degrades as more sampling steps are allocated at test time, limiting their effectiveness for any-step video diffusion.
  We argue that this limitation arises because consistency distillation replaces the original probability-flow ODE trajectory with a consistency-sampling trajectory, weakening the desirable test-time scaling behavior of ODE sampling.
  To address this limitation, we introduce \textbf{AnyFlow}, the first any-step video diffusion distillation framework based on flow maps.
  Instead of distilling a model for only a few fixed sampling steps, AnyFlow optimizes the full ODE sampling trajectory.
  To this end, we shift the distillation target from endpoint consistency mapping ($\mathbf{z}_t\!\to\!\mathbf{z}_0$) to flow-map transition learning ($\mathbf{z}_t\!\to\!\mathbf{z}_r$) over arbitrary time intervals.
  We further propose \textbf{Flow Map Backward Simulation}, which decomposes a full Euler rollout into shortcut flow-map transitions, enabling efficient on-policy distillation that reduces test-time errors (\ie discretization error in few-step sampling and exposure bias in causal generation).
  Extensive experiments across both bidirectional and causal architectures, at scales ranging from 1.3B to 14B parameters, demonstrate that AnyFlow achieves performance comparable to or better than consistency-based counterparts in the few-step regime, while supporting flexible and scalable sampling under varying step budgets. In the causal text-to-video setting, AnyFlow-FAR reaches a VBench score of 84.05 at 4 NFEs and further improves to 84.41 at 32 NFEs, surpassing Krea-Realtime-14B (83.25 at 4 NFEs). For image-to-video generation, AnyFlow-FAR achieves a VBench-I2V score of 87.87 at 4 NFEs, comparable to Wan2.1-I2V-14B using 50$\times$2 NFEs (87.71). In the 14B bidirectional text-to-video setting, AnyFlow reaches 84.04 at 4 NFEs, outperforming rCM-14B (83.73 at 4 NFEs).
  Code is released at \url{https://github.com/NVLabs/AnyFlow}.
\end{abstract}

\begin{document}
\maketitle

\section{Introduction}
\label{sec:intro}

Video diffusion models have recently achieved remarkable generation quality at scale~\cite{wan2025wan,agarwal2025cosmos,kong2024hunyuanvideo,yang2024cogvideox,HaCohen2024LTXVideo}, but most pipelines remain tied to fixed inference-step budgets. In practice, users require flexible generation: rapid previews for quick iteration and higher-fidelity outputs for final delivery. This motivates the need for \emph{any-step} models that can trade latency for quality at test time without retraining.

\begin{figure*}[!tb]
  \centering
  \begin{subfigure}[b]{\linewidth}
    \centering
    \includegraphics[width=\linewidth]{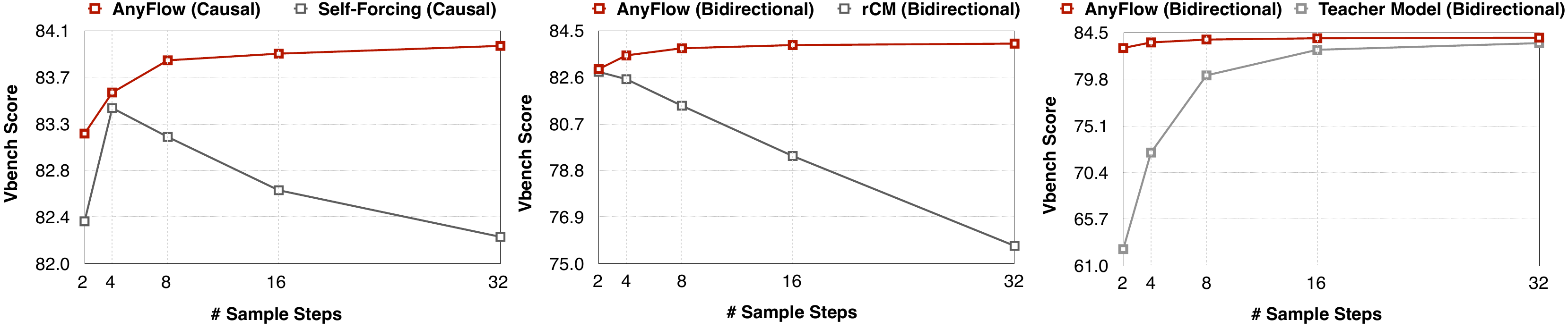}
    \caption{Quantitative comparison of AnyFlow test-time scaling against consistency-based methods and the teacher model.}
  \end{subfigure}
  \hfill
  \begin{subfigure}[b]{\linewidth}
    \centering
    \animategraphics[width=\linewidth]{8}{figures/videosrc/teaser/}{00000}{00020}
    \caption{Qualitative comparison of AnyFlow test-time scaling against consistency-based methods and the teacher model.}
  \end{subfigure}
  \caption{\textbf{Test-Time Scaling of AnyFlow.} Compared to consistency-based methods (\eg rCM~\cite{zheng2025large} for bidirectional video diffusion and Self-Forcing~\cite{huang2025selfforcing} for causal video diffusion), AnyFlow achieves strong performance in the few-step regime and scales effectively with increased sample steps. Compared to the teacher model, AnyFlow preserves test-time scaling ability while significantly improving the full trajectory. Readers can \textcolor{magenta}{click and play} the video clips in this figure using Adobe Acrobat.}
  \label{fig:teaser}
\end{figure*}

\begin{figure}[!tb]
  \centering
  \includegraphics[width=\linewidth]{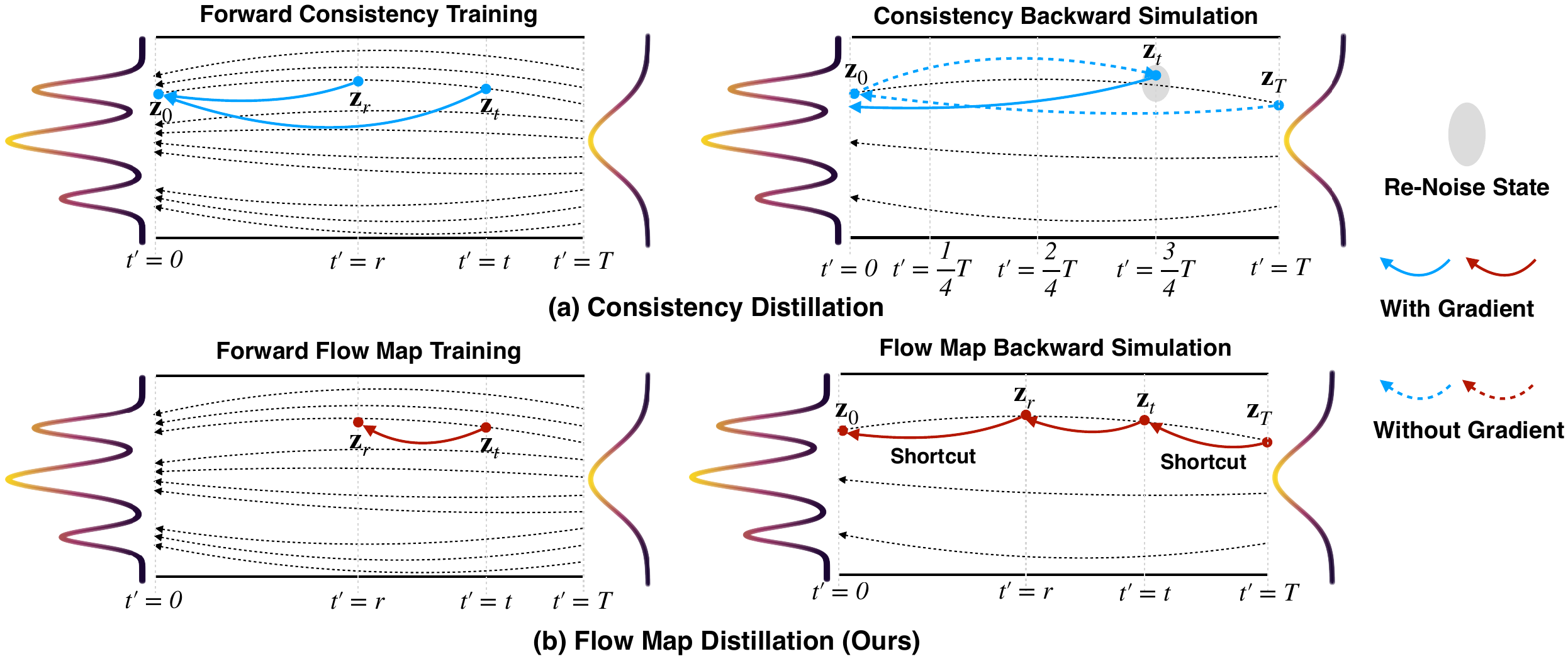}
  \caption{\textbf{Comparison of Distillation Paradigms.} (a) Consistency distillation learns a fixed-point mapping from $\mathbf{z}_t$ to $\mathbf{z}_0$. Its backward simulation requires re-noising intermediate states, which causes trajectory drift under multi-step sampling. (b) Flow map distillation learns transitions from $\mathbf{z}_t$ to $\mathbf{z}_r$. Its backward simulation decomposes Euler sampling trajectories into shortcut segments and enables efficient simulation of transitions across different step sizes.
  }
  \label{fig:paradigm_comp}
\end{figure}

Existing few-step video distillation methods are predominantly built on consistency models~\cite{song2023consistency,song2024improved,lu2024simplifying,zheng2025large,huang2025selfforcing}. While effective under very small sampling budgets, they do not provide robust any-step behavior. As illustrated in \cref{fig:teaser}, representative methods (\eg rCM~\cite{zheng2025large} for bidirectional models and Self-Forcing~\cite{huang2025selfforcing} for causal models) often degrade as the number of sampling steps increases, instead of improving with additional computation. The limitation is structural: consistency formulations emphasize fixed-point mappings to $\mathbf{z}_0$. During multi-step sampling, repeatedly re-noising intermediate states introduces cumulative bias, causing trajectories to drift away from the target PF-ODE path~\cite{sabour2025align} rather than being progressively refined, as shown in \cref{fig:paradigm_comp}(a).

To address this gap, we propose \textbf{AnyFlow}, the first any-step video diffusion distillation framework based on a two-time flow map formulation~\cite{boffi2024flow,sabour2025align,geng2025mean}. Flow maps generalize endpoint consistency by learning transitions between arbitrary time pairs, \ie $\mathbf{z}_t\!\to\!\mathbf{z}_r$ rather than only $\mathbf{z}_t\!\to\!\mathbf{z}_0$ (as shown in \cref{fig:paradigm_comp}(b)), which naturally supports variable step sizes and inference budgets.

Specifically, AnyFlow contains two complementary stages. In the first stage, we develop an improved forward flow map training recipe to convert pretrained video diffusion models into flow map models, providing a strong initialization for any-step generation. However, forward flow map training alone cannot fully remove test-time errors: discretization error remains pronounced under low-NFE sampling, and exposure bias is still severe in causal generation. In the second stage, we introduce on-policy flow map distillation to optimize reverse divergence on model rollouts to mitigate test-time errors. The core design is flow map backward simulation, which replaces expensive full-trajectory simulation with shortcut decomposition, enabling efficient training of intermediate transitions over different time ranges. By combining forward flow map training with reverse-divergence supervision during on-policy distillation, AnyFlow achieves strong few-step quality while continuing to improve with more sampling steps. Moreover, because AnyFlow preserves the fine-grained flow field, the distilled model can be further adapted to downstream datasets through continued fine-tuning.

We validate AnyFlow on both bidirectional and causal video diffusion models, ranging from 1.3B to 14B parameters. Across these settings, \textbf{AnyFlow matches or surpasses consistency-based counterparts even with few steps and continues to improve as the number of sampling steps increases}. In the causal setting, a single AnyFlow model jointly supports text-to-video, image-to-video, and video-to-video generation. For text-to-video, AnyFlow-FAR reaches 84.05 at 4 NFEs and further improves to 84.41 at 32 NFEs on the 14B model, surpassing Krea-Realtime-14B (83.25 at 4 NFEs). For image-to-video, AnyFlow-FAR achieves an 87.87 VBench-I2V score at 4 NFEs, comparable to Wan2.1-I2V-14B using 50$\times$2 NFEs (87.71). In the 14B bidirectional text-to-video setting, AnyFlow reaches 84.04 at 4 NFEs, outperforming rCM-14B (83.73 at 4 NFEs). Our contributions are summarized as follows:
\begin{itemize}
    \item We introduce AnyFlow, the first any-step video diffusion distillation framework built on flow maps, enabling a single model to support arbitrary inference budgets.
    \item We propose flow map backward simulation for on-policy flow map distillation to mitigate test-time errors (\ie discretization error and exposure bias) through efficient trajectory decomposition.
    \item We validate AnyFlow across architectures and model scales, establishing it as a scalable solution for high-quality, any-step video generation.
\end{itemize}

\section{Related Work}
\label{sec:related_work}

\subsection{Consistency Models}
Consistency Models (CMs) \cite{song2023consistency} accelerate diffusion sampling by directly learning a mapping from $\mathbf{z}_t$ to $\mathbf{z}_0$ along the probability-flow ODE (PF-ODE) defined by a teacher model. For multi-step generation, CMs repeatedly inject noise into intermediate states and perform iterative denoising. Subsequent studies improve CM training stability through annealed time schedules \cite{song2024improved, geng2024consistency} and segmented consistency objectives \cite{wang2024phased, ren2024hyper, lee2024truncated}. sCM \cite{lu2024simplifying} further simplifies these design choices and provides a strong practical baseline. Building on sCM, rCM \cite{zheng2025large} introduces score distillation as a regularizer, achieving strong performance in both image and video diffusion distillation. In contrast to prior CM-based approaches, we explore flow map distillation to enable any-step video generation.

\subsection{Flow Map Models}
Flow Map Models~\cite{boffi2024flow, sabour2025align, kim2024consistency} offer a unified perspective on diffusion modeling and diffusion distillation by learning a generalized transition operator $\mathbf{f}_\theta(\mathbf{z}_t, t, r)$ between arbitrary points along PF-ODE trajectories. From this perspective, the formulation recovers consistency modeling~\cite{song2023consistency} when $r=0$ and reduces to standard flow matching~\cite{liu2022flow, lipman2022flow} when $t=r$. However, training flow map models is challenging because it requires accurate transition learning across arbitrary time pairs. MeanFlow~\cite{geng2025mean} is a representative approach that connects instantaneous and average velocities, but it relies on Jacobian-vector products (JVPs), which are difficult to scale under Fully Sharded Data Parallel (FSDP). To overcome this limitation, subsequent studies either approximate JVP terms numerically~\cite{wang2025transition} or derive JVP-free algebraic formulations~\cite{luo2025soflow, guo2025splitmeanflow}. Building on these advances, AnyFlow leverages flow map distillation for scalable any-step video diffusion, enabling efficient and flexible inference for high-fidelity generation.

\subsection{Video Diffusion Distillation}
As shown in \cref{tab:paradigm_comp}, recent video diffusion distillation methods typically follow a two-stage pipeline: (1) forward training for initialization and (2) on-policy distillation for refinement. The first stage improves optimization stability, while the second reduces rollout drift during few-step inference. For example, rCM~\cite{zheng2025large} initializes from sCM~\cite{lu2024simplifying} and then applies consistency backward simulation~\cite{yin2024improved} with distribution matching~\cite{yin2024one}. In causal video generation, Self-Forcing~\cite{huang2025selfforcing} adopts data-free consistency ODE initialization~\cite{yin2025causvid} before performing a similar on-policy refinement stage. Our method follows the same high-level pipeline but replaces consistency modeling with
a flow map formulation, enabling stronger any-step performance.

A concurrent work, TMD~\cite{nie2026transition}, also studies a flow map formulation for bidirectional video diffusion distillation. The main difference lies in how efficient rollout is achieved. TMD improves rollout efficiency from an architectural perspective by sharing the backbone and introducing an additional flow head. In contrast, our method improves simulation efficiency from the flow-trajectory perspective by decomposing a full trajectory into shortcut transitions based on the composition property of flow maps. As a result, our approach naturally supports arbitrary step budgets and generalizes to both bidirectional and causal architectures.

\myPara{Causal Video Diffusion.} A growing line of work studies causal video diffusion models~\cite{yang2025longlive, jinpyramidal,zhang2025framepack,songhistory}. These models often suffer from exposure bias, which leads to error accumulation during autoregressive generation. Self-Forcing~\cite{huang2025selfforcing} mitigates this issue through on-policy distillation, but its consistency-based formulation is mainly tailored to few-step settings for rollout efficiency. Another line of work~\cite{guo2025end} focuses more on multi-step models by explicitly modeling test-time errors during pretraining, but it is not directly designed for efficient few-step sampling. By contrast, our method starts from a flow map formulation and uses flow map backward simulation to support both few-step and multi-step causal video generation within a unified framework.

\subsection{On-Policy Distillation}
On-policy distillation mitigates test-time errors  by training the student on its own generated trajectories and supervising them with a stronger teacher. This approach has demonstrated benefits in both large language models \cite{lu2025onpolicy, zhao2026self, ye2026policy} and diffusion models \cite{yin2024one, huang2025selfforcing, yin2024improved}.

For diffusion models, on-policy distillation is typically implemented via Distribution Matching Distillation (DMD)~\cite{yin2024one, yin2024improved} or adversarial distillation~\cite{lindiffusion, lin2025autoregressive}. Self-Forcing~\cite{huang2025selfforcing} combines self-rollouts with distribution matching against a bidirectional teacher, while APT2~\cite{lin2025autoregressive} evaluates one-step video rollouts using a discriminator. Our method follows the same on-policy distillation paradigm but introduces flow map backward simulation, specifically designed for the flow map formulation, enabling any-step video generation. \begin{table*}[!tb]
\centering
\resizebox{\linewidth}{!}{\begin{tabular}{l| c c| c c c}
\toprule
\textbf{Method} & \textbf{Forward Training} & \textbf{On-Policy Distillation} & \textbf{Causal} & \textbf{Bidirectional} & \textbf{Any-Step} \\
\midrule
APT1\&2~\cite{lindiffusion,lin2025autoregressive} & Consistency Training & One-Step Backward Simulation + GAN Loss & \cmark & \cmark & \xmark \\
rCM~\cite{zheng2025large} & Consistency Training & Consistency Backward Simulation + DMD Loss & \xmark & \cmark & \xmark \\

Self-Forcing~\cite{huang2025selfforcing} & Consistency ODE Init & Consistency Backward Simulation + DMD Loss & \cmark & \xmark & \xmark \\

\rowcolor{mitblue}AnyFlow & Flow Map Training & Flow Map Backward Simulation + DMD Loss & \cmark & \cmark & \cmark \\
\bottomrule
\end{tabular}}
\caption{\textbf{Comparison of Video Diffusion Distillation Methods.} Unlike prior consistency-based methods, AnyFlow is built on flow map modeling and supports both causal and bidirectional any-step generation.
}
\label{tab:paradigm_comp}
\end{table*}
\section{Preliminary}
\label{sec:preliminary}

\subsection{Flow Map Formulation}

We first introduce the basic flow map formalism used in AnyFlow.
Let $\mathbf{z}_t$ denote the latent state at continuous time $t\in[0,1]$ under the probability-flow ODE (PF-ODE):
\begin{equation}
  \frac{d\mathbf{z}_t}{dt} = \mathbf{v}(\mathbf{z}_t, t),
\end{equation}
where $\mathbf{v}$ is the velocity field.

Define the exact flow map $\Phi_{r\leftarrow t}$ of this ODE as the operator that transports states from time $t$ to time $r$, \ie $\Phi_{r\leftarrow t}(\mathbf{z}_t)=\mathbf{z}_r$ for $1\ge t\ge r\ge 0$.
It satisfies two standard properties: identity $\Phi_{t\leftarrow t}(\mathbf{z})=\mathbf{z}$ and composition $\Phi_{q\leftarrow r}\circ\Phi_{r\leftarrow t}=\Phi_{q\leftarrow t}$ for $t\ge r\ge q$.

In practice, a neural flow map model learns an approximation
\begin{equation}
  \mathbf{f}_\theta(\mathbf{z}_t, t, r) \approx \mathbf{z}_r, \quad 1 \ge t > r \ge 0,
\end{equation}
with boundary condition $\mathbf{f}_\theta(\mathbf{z}_t,t,t)=\mathbf{z}_t$.
Compared with endpoint-only mappings, this parameterization models transitions between arbitrary time pairs, which naturally supports variable step sizes and any-step inference.

\begin{table}[!tb]
    \centering
    \resizebox{\linewidth}{!}{
    \begin{tabular}{c|c|ccc|ccc}
        \toprule
        \multirow{2}{*}{\textbf{Method}} & \multirow{2}{*}{\textbf{NFEs}} & \multicolumn{3}{c|}{\textbf{Bidirectional Video Diffusion}} & \multicolumn{3}{c}{\textbf{Causal Video Diffusion}} \\
        \cmidrule(lr){3-5} \cmidrule(lr){6-8}
        & & \textbf{Quality} & \textbf{Semantic} & \textbf{Overall} & \textbf{Quality} & \textbf{Semantic} & \textbf{Overall} \\
        \midrule\midrule
        \multicolumn{8}{c}{\textbf{Forward Training}}\\
        \midrule\midrule
        \multirow{2}{*}{Flow Matching Training} & 4$\times$2 & 77.82 & 61.91 & 74.64 & 79.05 & 67.79 & 76.80 \\
        & 32$\times$2 & \textbf{85.48} & \textbf{77.24} & \textbf{83.83} & 85.21 & \textbf{76.65} & \textbf{83.50} \\\midrule
        \multirow{2}{*}{Consistency ODE-Init~\cite{yin2025causvid,huang2025selfforcing}} & 4 & 83.01 & 70.13 & 80.44 & 76.99 & 61.88 & 73.97 \\
        & 32 & 84.78 & 75.15 & 82.86 & 79.73 & 68.81 & 77.55\\\midrule
        \rowcolor{mitblue} & 4 & \textbf{84.39} & \textbf{71.20} & \textbf{81.75} & \textbf{82.80} & \textbf{71.16} & \textbf{80.48} \\
        \rowcolor{mitblue}\multirow{-2}{*}{Flow Map Training}& 32 & 85.35 & 75.63 & 83.40 & \textbf{85.23} & 74.71 & 83.13 \\
        \midrule\midrule
        \multicolumn{8}{c}{\textbf{Forward Training + On-Policy Distillation}}\\
        \midrule\midrule
        Consistency ODE-Init + & 4 & 84.77 & 75.71 & 82.96 & 84.37 & 74.97 & 82.49 \\
        Consistency Backward Simulation~\cite{huang2025selfforcing} & 32 & 83.70 & 64.18 & 79.80 & 82.43 & 68.48 & 79.64 \\
        \midrule
        Flow Map Training + & 4 & \textbf{85.47} & 75.88 & \textbf{83.55} & 84.99 & 74.97  & 82.99 \\
        Consistency Backward Simulation & 32 & 85.15 & 74.19 & 82.96 & 85.67 & 74.78 & 83.49 \\\midrule
        \rowcolor{mitblue}Flow Map Training + & 4 & 85.24 & \textbf{76.41} & 83.48 & \textbf{85.60} & \textbf{75.30} & \textbf{83.54} \\
        \rowcolor{mitblue}Flow Map Backward Simulation (Ours) & 32 & \textbf{85.70} & \textbf{76.99} & \textbf{83.96} & \textbf{85.92} & \textbf{76.12} & \textbf{83.96} \\
        \bottomrule
        \end{tabular}}    \caption{\textbf{Quantitative ablation of key designs in AnyFlow.}
        Flow map training provides a stronger initialization for few-step sampling than flow matching training and consistency ODE-Init. However, forward training alone still suffers from test-time errors, which are further mitigated by the on-policy distillation stage. Among the evaluated designs, flow map backward simulation delivers the strongest performance in both few-step and multi-step settings.
      }
    \label{tab:ablation_anyflow}
\end{table}

\myPara{MeanFlow Objective.}
MeanFlow~\cite{geng2025mean} is a representative algorithm for training flow map models. It parameterizes the averaged transport velocity on $[r,t]$ and approximates the transition as $\mathbf{f}_\theta(\mathbf{z}_t,t,r)=\mathbf{z}_t-(t-r)\,\mathbf{u}_\theta(\mathbf{z}_t,r,t)$. Following this formulation, we optimize $\mathbf{u}_\theta$ by:
\begin{equation}
  \label{eq:meanflow}
  \mathcal{L}(\theta)
  = \mathbb{E}\left[\left\|\mathbf{u}_\theta(\mathbf{z}_t, r, t)-\operatorname{sg}\!\left(\mathbf{u}_{\text{tgt}}\right)\right\|_2^2\right],
\end{equation}
where $\mathbf{u}_{\text{tgt}}=\mathbf{v}(\mathbf{z}_t,t) - (t-r)\,\frac{d\mathbf{u}_\theta(\mathbf{z}_t,r,t)}{dt}$.
Here, $\operatorname{sg}(\cdot)$ denotes stop-gradient.

\myPara{Differential Derivation
Equation.} Computing the Jacobian-vector product (JVP) for the derivative term $\frac{d\mathbf{u}(\mathbf{z}_t, r, t)}{dt}$ is expensive and not fully compatible with Fully Sharded Data Parallel (FSDP). Transition Model~\cite{wang2025transition} proposes to approximate this derivative using the finite difference method:
\begin{equation}
    \label{eq:center_difference}
  \frac{d}{dt}\mathbf{u}(\mathbf{z}_t, r, t) \approx \frac{\mathbf{u}(\mathbf{z}_{t+\Delta t}, r, t+\Delta t) - \mathbf{u}(\mathbf{z}_{t-\Delta t}, r, t-\Delta t)}{2\Delta t},
\end{equation}
This approximation requires only two forward passes per training step and is compatible with FSDP.

\subsection{On-Policy Distillation}
\label{sec:on_policy_distillation}
On-policy distillation is designed to reduce train--test mismatch by training the student on its own rollout states while being guided by a strong teacher. In diffusion models, this paradigm is mainly instantiated as Distribution Matching Distillation (DMD). It typically consists of two key components: backward simulation and a distribution-matching objective.

\myPara{Backward Simulation.}
Given noise $\mathbf{z}$, the student first produces a sample through its inference trajectory,
$\hat{\mathbf{z}}_0 = \mathbf{f}_\theta(\mathbf{z})$.
This self-generated trajectory is then used to compute the distillation loss, which helps mitigate exposure bias and discretization error at test time.

\myPara{Distribution Matching Objective.}
We follow DMD-style reverse-KL training~\cite{yin2024one,yin2024improved}. First, DMD re-noises the self-rollout sample $\hat{\mathbf{z}}_0$ at $t\in[0,T]$ as $\mathbf{z}_t=(1-t)\hat{\mathbf{z}}_0+t\boldsymbol{\epsilon}$, where $\boldsymbol{\epsilon}\sim\mathcal{N}(\mathbf{0},\mathbf{I})$. It then estimates the DMD gradient as follows:
\begin{align}
  \label{eq:DMD_gradient}
  \nabla_\theta \mathcal{L}_{\text{DMD}}
  = -\mathbb{E}_{t,\mathbf{z}}\left[
    \big(s_{\text{real}}(\mathbf{z}_t,t)-s_{\text{fake}}(\mathbf{z}_t,t)\big)
    \frac{\partial \mathbf{f}_\theta(\mathbf{z})}{\partial \theta}
  \right],
\end{align}
where $s_{\text{real}}$ and $s_{\text{fake}}$ are the real and fake score functions, respectively.

\begin{figure*}[!tb]
  \centering
  \animategraphics[width=\linewidth]{8}{figures/videosrc/ablation_onpolicy/}{00000}{00020}
   \caption{\textbf{Qualitative Ablation of On-Policy Distillation.} (a) Forward flow map training alone still suffers from test-time errors, including discretization error and exposure bias. (b) On-policy flow map distillation optimizes reverse divergence on model self-rollouts to mitigate these test-time errors. Readers can \textcolor{magenta}{click and play} the video clips in this figure using Adobe Acrobat.}
  \label{fig:ablation_onpolicy}
\end{figure*}

\section{Method}
\label{sec:method}

\subsection{AnyFlow Motivation}
\label{sec:motivation}

As summarized in \cref{tab:paradigm_comp}, most recent few-step video distillation methods are built on consistency models. Although effective under very small sampling budgets, their performance often degrades as the number of sampling steps increases (\cref{fig:teaser}). The underlying reason is that consistency-style sampling repeatedly performs endpoint projection and re-noising, which gradually drives the trajectory away from the target PF-ODE path during multi-step inference.

\myPara{Motivation.} Our goal is to preserve the strong few-step efficiency of distilled video diffusion models while enabling quality to improve, rather than deteriorate, as more sampling steps are used. To this end, AnyFlow shifts the distillation target from fixed-point endpoint mapping to flow map transition learning, \ie $\mathbf{f}_\theta:(\mathbf{z}_t, t, r) \mapsto \mathbf{z}_r$. Instead of only predicting $\mathbf{z}_0$, the model learns transitions between arbitrary time pairs. This formulation naturally supports both small time gaps for stable local refinement and large time gaps for efficient long-range jumps, making it well suited for any-step generation.

\myPara{Challenge.} Forward flow map training alone (\eg MeanFlow~\cite{geng2025mean}) is insufficient for strong test-time performance in video generation. As shown in \cref{fig:ablation_onpolicy} and \cref{tab:ablation_anyflow}, it still suffers from test-time errors, especially discretization error in few-step sampling and exposure bias in causal generation. This motivates an additional on-policy distillation stage to correct rollout mismatch. However, designing such a stage for flow map models is non-trivial. Unlike consistency models, which can naturally reach $\mathbf{z}_0$ at intermediate steps and therefore allow KL gradients to be computed directly at endpoint states, flow map models require a new simulation strategy that can accommodate transitions between arbitrary time pairs while remaining efficient for training.

\begin{figure*}[!tb]
    \centering
    \includegraphics[width=\linewidth]{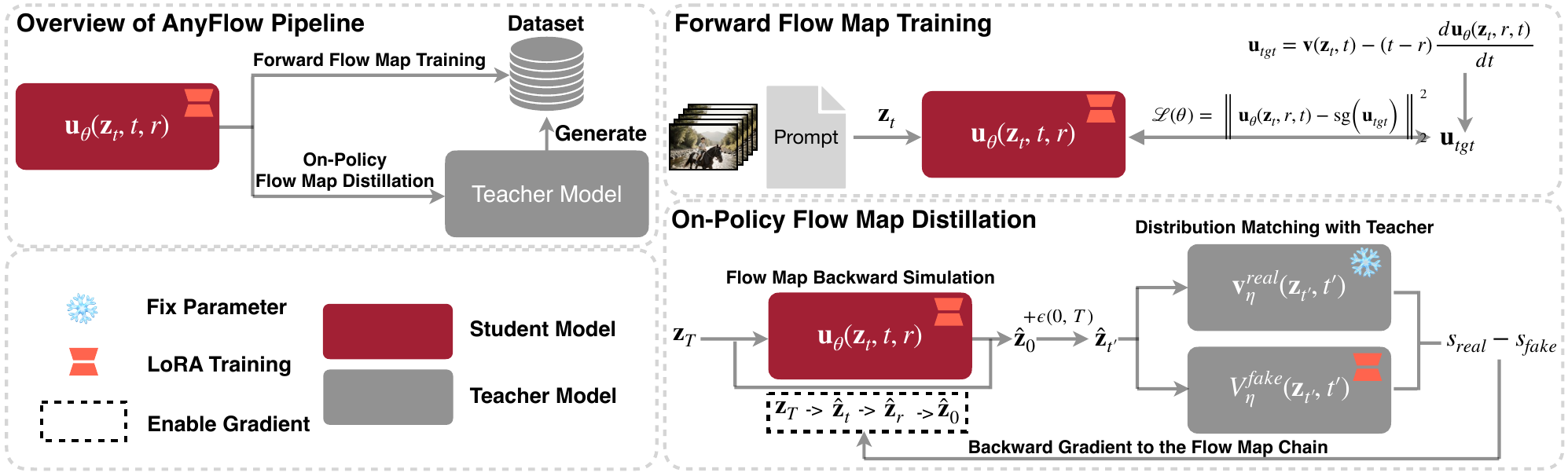}
    \caption{\textbf{Overview of the AnyFlow Pipeline.} AnyFlow enables any-step video generation by jointly learning forward flow map training from synthetic data and on-policy distillation with flow map backward simulation under teacher guidance.
  }
    \label{fig:pipeline}
\end{figure*}

\subsection{AnyFlow Pipeline}
\label{sec:pipeline}

\myPara{Overview.} As illustrated in \cref{fig:pipeline}, AnyFlow is trained in two complementary stages: forward flow map training (\cref{sec:flow_map}) and on-policy distillation (\cref{sec:on_policy}). In Stage 1, we fine-tune a pretrained video diffusion model on teacher-synthesized data to learn a stable transition operator $\mathbf{f}_\theta:(\mathbf{z}_t, t, r) \mapsto \mathbf{z}_r$, which provides a strong initialization for any-step generation. In Stage 2, starting from the converged Stage-1 checkpoint, we jointly optimize forward flow map training and DMD-based on-policy distillation. This stage uses flow map backward simulation to generate student rollouts and apply reverse-divergence supervision from a strong teacher, thereby reducing discretization error and exposure bias while preserving any-step capability.

\subsubsection{Flow Map Training}
\label{sec:flow_map}

\noindent We employ the MeanFlow objective (\cref{eq:meanflow}) alongside the differential derivation equation (\cref{eq:center_difference}) as our primary training framework. Because the original MeanFlow architecture is optimized for pretraining, directly adapting it to post-training for video diffusion models requires several additional design choices for strong performance. To this end, we introduce several modifications: interpolated timestep conditioning, time sampling, guidance-fused training, and adaptive loss reweighting.

\definecolor{commentcolor}{RGB}{84, 137, 137}
\definecolor{funccolor}{RGB}{201, 62, 116}
\newcommand{\mycomment}[1]{\textcolor{commentcolor}{\# #1}}
\newcommand{\func}[1]{\textcolor{funccolor}{\texttt{#1}}}

\setlength{\algomargin}{0pt}
\setlength{\AlCapHSkip}{0pt}
\SetAlgoCaptionLayout{raggedright}

\begin{figure}[!tb]
    \begin{minipage}{0.49\textwidth}
        \begin{algorithm}[H]
            \caption{AnyFlow: Forward Flow Map Training}
            \label{alg:flowmap}
            \SetInd{0.3em}{0.8em}
            \DontPrintSemicolon

            \mycomment{$\text{fn}(\mathbf{z}, c, t, r)$: model to predict $\mathbf{u}$} \;
            \mycomment{$\mathbf{x}$: training batch} \;
            \mycomment{c: text embedding} \;
            \mycomment{g: CFG scale} \;
            \vspace{1em}

            $t, r = \func{sample\_t\_r}() \quad\mycomment{\ t>r}$\;
            $\mathbf{e} = \func{randn\_like}(\mathbf{x})$\;
            \vspace{1em}
            
            $\mathbf{z} = (1 - t)*\mathbf{x} + t*\mathbf{e}$\;
            $\mathbf{v} = \mathbf{e} - \mathbf{x}$\;
            \vspace{1em}

            \mycomment{Guidance-Fused Training}\;
            $\mathbf{u}_c = \text{fn}(\mathbf{z}, c, t, r)$\;
            $\mathbf{u}_{\varnothing} = \text{fn}(\mathbf{z}, \varnothing, t, r)$\;
            $\mathbf{u} = \frac{1}{g}\left(\mathbf{u}_c - (1-g)*\func{stopgrad}(\mathbf{u}_{\varnothing})\right)$
            \vspace{1em}

            \mycomment{Differential Derivation Equation}\;
            $\mathbf{z}^{+} = \mathbf{z} + \epsilon\mathbf{v}$\;
            $\mathbf{z}^{-} = \mathbf{z} - \epsilon\mathbf{v}$\;

            $\frac{d\mathbf{u}}{dt} = \frac{\text{fn}(\mathbf{z}^+, c, t+\epsilon, r)-\text{fn}(\mathbf{z}^-, c, t-\epsilon, r)}{2\epsilon g}$\;
            $\mathbf{u}_{tgt} = \mathbf{v} - (t - r)\frac{d\mathbf{u}}{dt}$\;
            \vspace{1em}
            
            $loss = \func{metric}(\mathbf{u} - \func{stopgrad}(\mathbf{u}_{tgt}))$\;
        \end{algorithm}
    \end{minipage}
    \hfill
    \begin{minipage}{0.49\textwidth}
        \begin{algorithm}[H]
            \caption{AnyFlow: On-Policy Flow Map Distillation}
            \label{alg:onpolicy}
            \SetInd{0.3em}{0.8em}
            \DontPrintSemicolon

            \mycomment{$\text{fn}(\mathbf{z}, c, t, r)$: model to predict $\mathbf{u}$} \;
            \mycomment{$\mathbf{x}$: training batch} \;
            \mycomment{c: text embedding} \;
            
            \vspace{1em}

            \While{true}{
            \vspace{1em}
            \mycomment{Sample Inference Step}\;
        $s \sim \text{Uniform}(1, 2, \dots, T)$\;
        \vspace{1em}
        \mycomment{Sample Gradient Step}\;
        $t \sim \text{Uniform}(1, 2, \dots, s)$\;
        $r = t - T/s$
    
        \vspace{1em}
        \mycomment{Sample Initial Noise}\;
        $\mathbf{z}_T = \func{randn\_like}(\mathbf{x})$\;
        \vspace{1em}
        \mycomment{Flow Map Backward Simulation} \;
        $\mathbf{z}_t = \text{fn}(\mathbf{z}_{T}, c, T, t)$ \mycomment{With Grad}\;  
        
        $\mathbf{z}_r = \text{fn}(\mathbf{z}_{t}, c, t, r)$ \mycomment{With Grad}\;
        $\mathbf{z}_0 = \text{fn}(\mathbf{z}_{r}, c, r, 0)$ \mycomment{With Grad}\;
        
        \vspace{1em}
        Update $\theta$ via distribution matching loss using $\mathbf{z}_0$ and renoise at [0, T] \;
    }
        \end{algorithm}
    \end{minipage}
\end{figure}

\myPara{Interpolated Timestep Conditioning.} Converting a pretrained diffusion model into a flow map model requires introducing an additional timestep, $r$. A natural choice is to follow MeanFlow~\cite{geng2025mean} and use $\text{emb}(t) + \text{emb}'(t-r)$. Prior work such as TMD~\cite{nie2026transition} further applies a zero-initialized output projection to $\text{emb}'$ for a smooth start in post-training. However, we find this design unstable in our setting: it tends to produce timestep embeddings with much larger norms than the pretrained ones, leading to over-saturated generation results, as shown in \cref{fig:ablation_interpolated_time}.

To address this issue, we instead use an interpolated timestep conditioning scheme:
$$g \cdot \text{emb}(t) + (1-g) \cdot \text{emb}'(r),$$
where $\text{emb}'$ is initialized from the pretrained $\text{emb}$ and $g$ is fixed to 0.25 in all experiments. At the beginning of training, this formulation reduces to the pretrained timestep embedding in the boundary case $t=r$. As verified in \cref{fig:ablation_interpolated_time}, interpolated timestep conditioning keeps the embedding norm well aligned with the pretrained model and suppresses over-saturated results observed in zero-init timestep conditioning.

\myPara{Time Sampler.}
Following MeanFlow~\cite{geng2025mean}, we first sample $t$ and $r$ from a uniform distribution and then reorder them as $t=\max(t,r)$ and $r=\min(t,r)$. We further introduce a reweighting function $w(t)$ based on the sampled timestep $t$ to balance the training objective across different noise levels. Different choices of $w(t)$ are discussed in the \cref{fig:ablation_time_sampler}.

\myPara{Guidance-Fused Training.}
We also incorporate classifier-free guidance (CFG) into flow map training following MeanFlow~\cite{geng2025mean}, which allows CFG to be omitted at test time for faster inference. Unlike MeanFlow, which fuses the CFG objective into the target velocity field $\mathbf{u}_{\text{tgt}}$, we fuse it into the prediction to better align with the guidance scale defined by the pretrained diffusion model:
\begin{equation}
    \mathbf{u} = \frac{1}{g}\left(\mathbf{u}_c - (1-g)\operatorname{sg}(\mathbf{u}_{\varnothing})\right),
  \end{equation}
where $\operatorname{sg}(\cdot)$ denotes the stop-gradient operation and $\mathbf{u}_{\varnothing}$ is the model prediction with null conditioning.

\myPara{Adaptive Loss Reweighting.} To stabilize flow map training, we introduce an adaptive loss reweighting scheme. Unlike MeanFlow~\cite{geng2025mean}, which uses a weighting function $w = 1 / (\|\Delta\|_2^2 + c)$ for pre-training, where $\Delta$ denotes the regression loss and $c$ is a small constant, our approach is designed for post-training, where a reliable instantaneous velocity field has already been established at $t=r$. We use the loss at $t=r$ as a baseline to dynamically scale the loss for other time steps $t \neq r$. In each iteration, we sample 50\% of the batch from the boundary cases ($t=r$) and define the adaptive weight $w_{t,r}$ as:
\begin{equation}
  w_{t,r} =
  \begin{cases} 1, & t = r \\ \displaystyle \frac{\mu_{t=r}}{\|\Delta\|_2^2 + c}, & t \neq r
  \end{cases}
\end{equation}
where $\mu_{t=r} = \mathbb{E}_{t=r} [ \|\Delta\|_2^2 ]$ is the average regression loss over the boundary samples. This formulation aligns the loss magnitude at $t \neq r$ with the well-optimized field at $t=r$, thereby preserving the learned instantaneous velocity and preventing gradient instability.

\myPara{Discussion.} The full algorithm of our improved forward flow map training is summarized in \cref{alg:flowmap}. As shown in \cref{tab:ablation_anyflow}, compared with standard flow matching, flow map training achieves comparable multi-step performance while delivering significantly better few-step performance. Compared with Consistency ODE-Init, it achieves stronger multi-step performance while maintaining comparable few-step quality. Overall, flow map training provides a strong initialization for the subsequent on-policy distillation stage.

\begin{figure}[!tb]
  \centering
  \includegraphics[width=\linewidth]{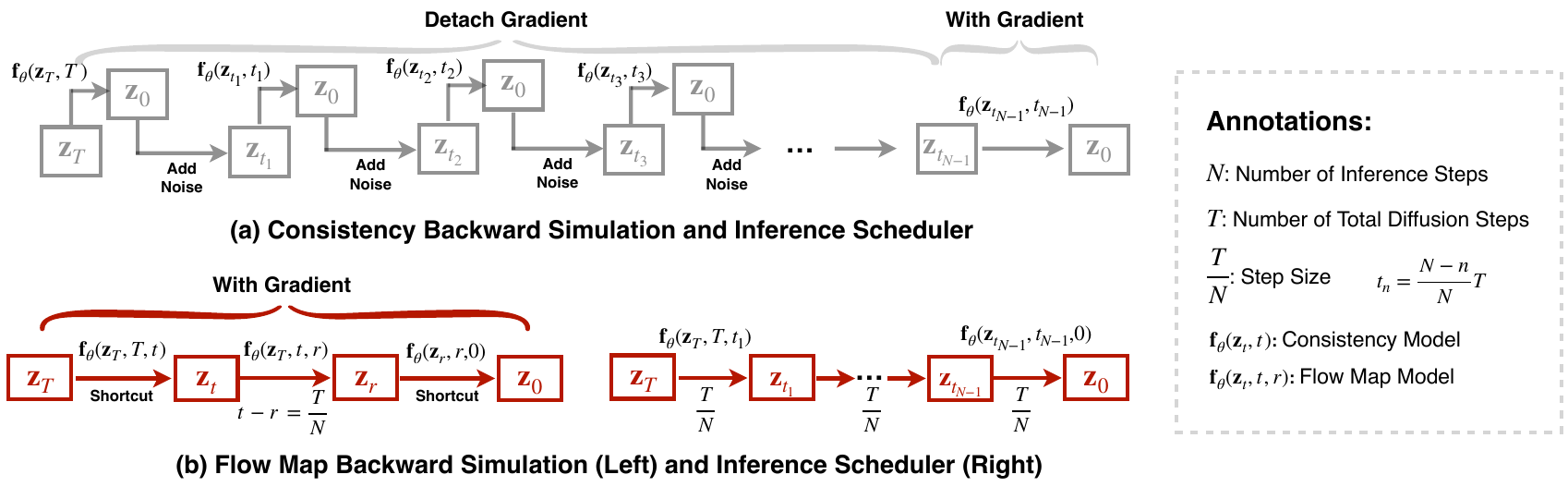}
  \caption{\textbf{Comparison of Backward Simulation Paradigms.} (a) Consistency backward simulation and sampling both follow consistency trajectories, where truncated gradients are used to reduce training cost of on-policy distillation. (b) Flow map backward simulation enables efficient simulation of Euler sampling trajectories. It decomposes the full rollout trajectory into shortcut segments to reduce the training cost of on-policy distillation.
  }
  \label{fig:comp_simulation}
\end{figure}

\begin{figure*}[!tb]
    \centering
    \animategraphics[width=\linewidth]{8}{figures/videosrc/ablation_anyflow_simulation/}{00000}{00020}
      \caption{\textbf{Qualitative Ablation of Backward Simulation.} (a) Consistency backward simulation exhibits over-smoothed textures and degraded motion as more sampling steps are allocated. (b) Flow map backward simulation produces a straighter trajectory, enabling test-time scaling while maintaining strong few-step quality. Readers can \textcolor{magenta}{click and play} the video clips in this figure using Adobe Acrobat.}
\label{fig:ablation_backward_simulation}
\end{figure*}

\subsubsection{On-Policy Flow Map Distillation}
\label{sec:on_policy}

\noindent Although forward flow map training provides a strong initialization for any-step generation, it still suffers from test-time errors, especially discretization error at low NFEs and exposure bias in causal generation, as shown in \cref{fig:ablation_onpolicy}. To correct rollout drift, we further introduce on-policy distillation with teacher guidance. Following prior work~\cite{huang2025selfforcing, zheng2025large}, we instantiate on-policy diffusion distillation with distribution matching distillation (DMD), which requires the student to perform a self-rollout (\ie backward simulation) to $\mathbf{z}_0$ before re-noising in order to compute the Kullback--Leibler (KL) gradient. Below, we describe our design of backward simulation for flow map
models.

\myPara{Limitations of Consistency Backward Simulation.}
Existing backward simulation methods for DMD~\cite{huang2025selfforcing, zheng2025large} largely follow the logic of consistency models, as illustrated in \cref{fig:comp_simulation}(a). This design is convenient for on-policy distillation because the consistency scheduler can reach the $\mathbf{z}_0$ state at intermediate steps, making endpoint-based supervision straightforward. In practice, gradients are often truncated before the target gradient step to avoid the memory cost and instability of backpropagating through a long rollout chain. In inference time, the same consistency sampler is then reused.

However, unlike ODE trajectories learned through flow-matching pretraining, consistency samplers require additional re-noising to obtain intermediate states during multi-step sampling. These re-noised states deviate from the base PF-ODE trajectory, which limits generalization beyond the simulation step used during simulation. Moreover, simulating arbitrary step counts with consistency backward simulation is computationally expensive because it requires rolling out the full trajectory.

\myPara{Flow Map Backward Simulation.}
Unlike consistency backward simulation, our design directly exploits the learned transition capability of flow map models. Since a flow map model is able to predict transitions between arbitrary time pairs, it can naturally shortcut long rollout trajectories instead of explicitly simulating every intermediate step, based on the composition property of flow maps:
\begin{equation}
  \mathbf{f}_{\theta}(\mathbf{z}_t,t,q) \approx \mathbf{f}_{\theta}\big(\mathbf{f}_{\theta}(\mathbf{z}_t,t,r),r,q\big), \quad t>r>q.
\end{equation}

Concretely, for a target sampling budget of $N$ steps, we first choose an intermediate timestep $t$ along the sampled trajectory and then set the next timestep $r$ such that $t-r = \frac{T}{N}$. As illustrated in \cref{fig:comp_simulation}, this decomposes a rollout trajectory from $T$ to $0$ into three segments: $T \to t$, $t \to r$, and $r \to 0$. The first and last segments, $T \to t$ and $r \to 0$, are handled by shortcut transitions of the learned flow map, while $t \to r$ is the target transition along the sampled trajectory. Once the KL gradient is computed at $z_0$, it is backpropagated through the whole chain. By varying $N$ during training, we can efficiently simulate different inference step budgets with the same computation cost.

After on-policy distillation, we can simply use a standard Euler scheduler for sampling. Because flow map backward simulation trains transitions over different time intervals, the same model naturally supports different inference budgets at test time.

\begin{figure}[!tb]
  \centering
  \includegraphics[width=\linewidth]{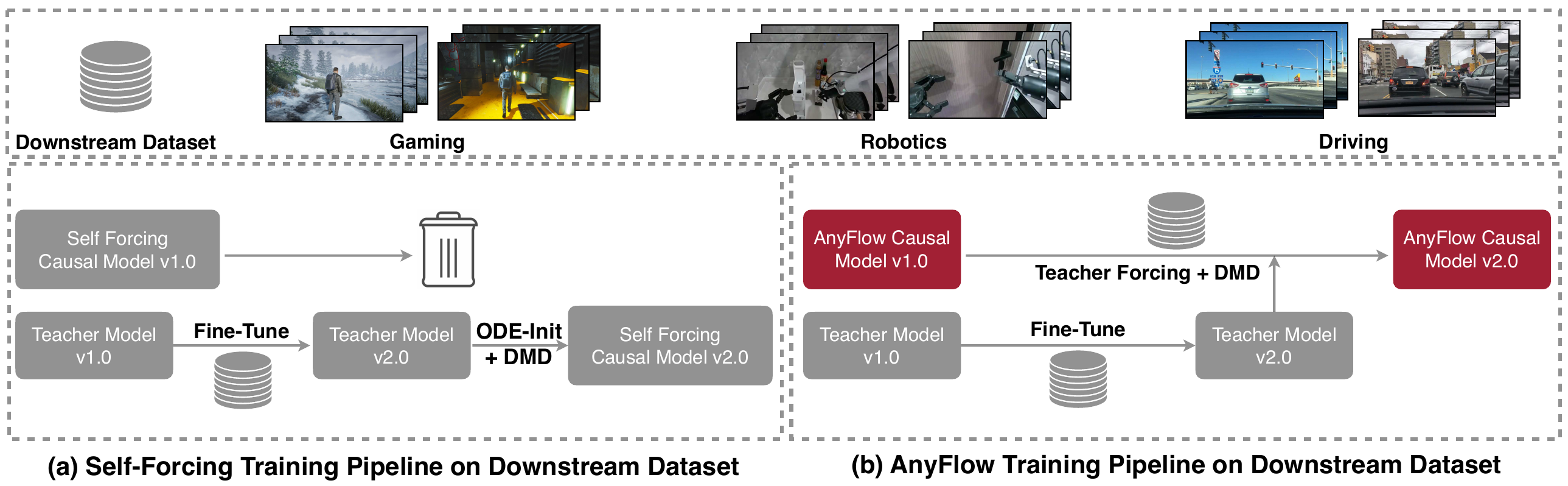}
  \caption{\textbf{Illustration of the AnyFlow Fine-Tuning Pipeline for Downstream Applications.} Unlike self-forcing pretrained causal models that are difficult to adapt to new downstream datasets, AnyFlow supports continued training. This capability bypasses the complexities of retraining a causal generator.
  }
  \label{fig:downstream_pipeline}
\end{figure}

\begin{figure}[!tb]
  \centering
  \includegraphics[width=\linewidth]{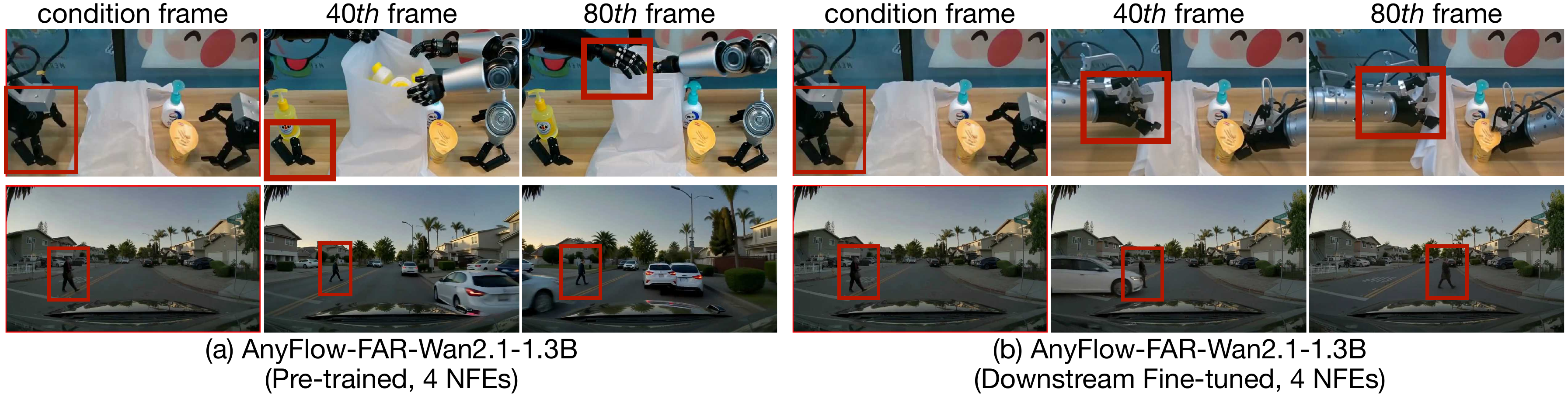}
  \caption{\textbf{Visualization of the AnyFlow Fine-Tuning for Downstream Applications.} While the pre-trained AnyFlow-FAR-Wan2.1-1.3B struggles with identity preservation (\eg robot arm type) and trajectory accuracy (\eg moving pedestrians) in specialized domains, AnyFlow supports continued training on downstream datasets, allowing for efficient application-specific tuning without full retraining.
  }
  \label{fig:downstream_result}
\end{figure}

\myPara{Discussion.} The full algorithm of our on-policy distillation stage is shown in \cref{alg:onpolicy}. As shown in \cref{tab:ablation_anyflow}, Consistency ODE-Init combined with consistency backward simulation exhibits the weakest test-time scaling, especially at 32 NFEs. Replacing the forward training stage with flow map training narrows this gap, but noticeable degradation still remains under multi-step sampling. By further adopting our flow map backward simulation, both few-step and multi-step performance are improved through on-policy distillation. As shown in \cref{fig:ablation_backward_simulation}, flow map backward simulation produces a straighter trajectory, which leads to more consistent improvement as the sampling budget increases, whereas consistency-based simulation degrades at 32 NFEs.

\subsection{AnyFlow Application}
\myPara{AnyFlow for Bidirectional Video Diffusion.}
For bidirectional video diffusion, we simply follow the established AnyFlow pipeline, as detailed in \cref{alg:flowmap} and \cref{alg:onpolicy}.

\myPara{AnyFlow for Causal Video Diffusion.}
For causal video diffusion, we adopt the FAR~\cite{gu2025long} training pipeline, which uses context compression with asymmetric patchify kernels to encode videos efficiently. Specifically, we keep three full-token chunks with the standard patchify kernel size of 2, while compressing the remaining chunks with a larger kernel size of 4. This design substantially reduces teacher-forcing training cost and KV-cache size during sampling.
To jointly support Image-to-Video (I2V) and Text-to-Video (T2V), we use a non-uniform chunk partition: the first chunk has size 1 for precise first-frame conditioning, and subsequent chunks use size 3 to balance motion modeling and throughput.
The training objective remains the same as in the bidirectional setting, and we additionally cache and reuse KV states during rollouts to improve simulation efficiency for causal video diffusion models.

\myPara{Continued Training on Downstream Datasets.}
A practical advantage of AnyFlow is that it preserves the pretrained model's instantaneous flow field, making the distilled model compatible with continued training on downstream datasets. As illustrated in \cref{fig:downstream_pipeline}, AnyFlow enables continued training from a pretrained checkpoint, unlike Self-Forcing, where further training of the distilled model is difficult.

This property is especially useful in specialized domains whose motion patterns and visual statistics differ from those of the original training corpus. As shown in \cref{fig:downstream_result}, the base distilled model, AnyFlow-FAR-Wan2.1-1.3B, still struggles with identity preservation in robotics videos and trajectory accuracy in driving scenes. After continued training on downstream data, these errors are substantially reduced.

\begin{table*}[!tb]
\centering
\resizebox{\linewidth}{!}{\begin{tabular}{l c c c c c c}
\toprule
\multirow{2}{*}{\textbf{Model}} & \multirow{2}{*}{\textbf{\#Params}} & \multirow{2}{*}{\textbf{Resolution}} & \multirow{2}{*}{\textbf{NFEs}} & \multicolumn{3}{c}{\textbf{Evaluation Scores $\uparrow$}} \\
\cmidrule(lr){5-7}
& & & & \textbf{Quality} & \textbf{Semantic} & \textbf{Total} \\
\midrule
\multicolumn{7}{l}{\textit{\textbf{Bidirectional Video Diffusion Models}}} \\
\midrule
LTX-Video~\tablecite{HaCohen2024LTXVideo} & 1.9B & $512\times768$ & 40$\times2$ & 82.30 & 70.79 & 80.00 \\
CogVideoX-2B~\tablecite{yang2024cogvideox} & 2B & $480\times720$ & 50$\times2$ & 82.48 & 77.81 & 81.55 \\
HunyuanVideo~\tablecite{kong2024hunyuanvideo} & 13B & $720\times1280$ & 50$\times2$ & 85.09 & 76.88 & 83.24 \\
\midrule
$^\dag$Wan2.1-T2V-1.3B~\tablecite{wan2025wan} & 1.3B & $480\times832$ & 50$\times 2$ & 84.99 & 76.23 & 83.24 \\
$^\dag$rCM-Wan2.1-T2V-1.3B~\tablecite{zheng2025large} & 1.3B & $480\times832$ & 4 & 84.71 & 73.74 & 82.51 \\
\rowcolor{mitblue}$^\dag$AnyFlow-Wan2.1-T2V-1.3B & 1.3B & $480\times832$ & 4 & 85.24 & 76.41 & 83.48 \\
\rowcolor{mitblue}$^\dag$AnyFlow-Wan2.1-T2V-1.3B & 1.3B & $480\times832$ & 32 & \textbf{85.70} & \textbf{76.99} & \textbf{83.96} \\
\midrule
$^\dag$Wan2.1-T2V-14B~\tablecite{wan2025wan} & 14B & $480\times832$ & 50$\times2$ & 85.77 & 75.58 & 83.74\\
$^\dag$rCM-Wan2.1-T2V-14B~\tablecite{zheng2025large} & 14B & $480\times832$ & 4 & 85.47 & 76.72 & 83.73\\
\rowcolor{mitblue}$^\dag$AnyFlow-Wan2.1-T2V-14B & 14B & $480\times832$ & 4 & 85.70 & 77.38 & 84.04 \\
\rowcolor{mitblue}$^\dag$AnyFlow-Wan2.1-T2V-14B & 14B & $480\times832$ & 32 & \textbf{85.76} & \textbf{77.44} & \textbf{84.10} \\
\midrule
\multicolumn{7}{l}{\textit{\textbf{Causal Video Diffusion Models}}} \\
\midrule
MAGI-1~\tablecite{teng2025magi} & 4.5B & $720\times720$ & 8 & 80.98 & 65.68 & 77.92 \\
SkyReels-V2~\tablecite{chen2025skyreels} & 1.3B & $540\times960$ & 50$\times2$ & 84.70 & 74.53 & 82.67 \\\midrule
$^\dag$Self-Forcing~\tablecite{huang2025selfforcing} & 1.3B & $480\times832$ & 4 & 85.23 & 76.01 & 83.39 \\
\rowcolor{mitblue}$^\dag$AnyFlow-FAR-Wan2.1-1.3B & 1.3B & $480\times832$ & 4 & 85.60 & 75.30 & 83.54 \\
\rowcolor{mitblue}$^\dag$AnyFlow-FAR-Wan2.1-1.3B & 1.3B & $480\times832$ & 32 & \textbf{85.92} & \textbf{76.12} & \textbf{83.96}\\\midrule
$^\dag$LightX2V-Wan2.1-14B-CausVid~\tablecite{lightx2v} & 14B & $480\times832$ & 9 & 85.29 & 75.96 & 83.42 \\
$^\dag$FastVideo-CausalWan2.2-A14B-Preview~\tablecite{causalwan2025} & 14B & $480\times832$ & 8 & 84.28 & \textbf{78.49} & 83.12 \\
$^\dag$Krea-Realtime-Wan2.1-14B~\tablecite{krea_realtime_14b} & 14B & $480\times832$ & 4 & 84.80 & 77.07 & 83.25 \\
\rowcolor{mitblue}$^\dag$AnyFlow-FAR-Wan2.1-14B & 14B & $480\times832$ & 4 & 85.82 & 76.97 &  84.05 \\
\rowcolor{mitblue}$^\dag$AnyFlow-FAR-Wan2.1-14B & 14B & $480\times832$ & 32 & \textbf{86.12} & 77.55 & \textbf{84.41} \\\midrule
\bottomrule
\end{tabular}}
\caption{\textbf{Text-to-Video Evaluation on VBench.} $\dag$ indicates that we re-evaluate model performance using the official VBench augmented prompts under the same setting. Results for all other models are taken directly from their respective papers.}
\label{tab:t2v_comparison}
\end{table*}

\begin{table*}[!tb]
\centering
\resizebox{\linewidth}{!}{\begin{tabular}{l c c c c c c}
\toprule
\textbf{Model} & \textbf{\#Params} & \textbf{Resolution} & \textbf{NFEs} & \multicolumn{3}{c}{\textbf{Evaluation Scores $\uparrow$}} \\
\cmidrule(lr){5-7}
& & & & \textbf{Quality} & \textbf{I2V} & \textbf{Total} \\
\midrule
CogVideoX-5B-I2V~\tablecite{yang2024cogvideox} & 5B & $480\times720$ & 50$\times2$ & 78.61 & 94.79 & 86.70 \\
HunyuanVideo-I2V~\tablecite{kong2024hunyuanvideo} & 13B & $720\times1280$ & 50$\times2$ & 78.54 & 95.10  & 86.82 \\
Step-Video-TI2V~\tablecite{huang2025step} & 30B & $540\times960$ & 50$\times2$ & 81.22 & 95.50 & 88.36 \\
MAGI-1~\tablecite{teng2025magi} & 24B & $720\times1280$ & 32$\times 2$ & 82.44 & 96.12 & 89.28\\\midrule
$^\dag$Wan2.1-I2V-14B~\tablecite{wan2025wan} & 14B & $480\times832$ & 50$\times 2$ & 80.30 & 95.12 & 87.71 \\
$^\dag$FastVideo-CausalWan2.2-A14B-Preview~\tablecite{causalwan2025} & 14B & $480\times832$ & 8 & 78.82 & 94.81 & 86.82 \\
\rowcolor{mitblue}$^\dag$AnyFlow-FAR-Wan2.1-14B & 14B & $480\times832$ & 4 & \textbf{80.39} & \textbf{95.35} & \textbf{87.87} \\
\bottomrule
\end{tabular}}
\caption{\textbf{Image-to-Video Evaluation on VBench-I2V.} $\dag$ indicates that we re-evaluate model performance under the same setting.}
\label{tab:i2v_comparison}
\end{table*}

\begin{figure*}[!tb]
    \centering
    \begin{subfigure}[b]{\linewidth}
        \centering
        \animategraphics[width=\linewidth]{8}{figures/videosrc/t2v14b_causal/}{00000}{00020}
        \caption{Text-to-video comparison of causal video diffusion model (14B).}
    \end{subfigure}
    \hfill 
    \begin{subfigure}[b]{\linewidth}
        \centering
        \animategraphics[width=\linewidth]{8}{figures/videosrc/t2v14b_bidirectional/}{00000}{00020}
        \caption{Text-to-video comparison of bidirectional video diffusion model (14B).}
    \end{subfigure}
    \hfill 
    \begin{subfigure}[b]{\linewidth}
        \centering
        \animategraphics[width=\linewidth]{8}{figures/videosrc/i2v14b/}{00000}{00020}
        \caption{Image-to-video comparison of video diffusion model (14B).}
    \end{subfigure}
    \caption{\textbf{Qualitative Comparison of AnyFlow and Baselines at the 14B Model Scale.} Readers can \textcolor{magenta}{click and play} the video clips in this figure using Adobe Acrobat.}
\label{fig:14b_compare}
\end{figure*}
 \section{Experiments}

\subsection{Implementation Details}
We implement AnyFlow on top of Wan2.1~\cite{wan2025wan} in the Diffusers framework~\cite{von-platen-etal-2022-diffusers}. We train the model on a synthetic dataset of 256K prompt--video pairs generated from Wan2.1-T2V-14B, where each sample contains up to 81 frames at a resolution of $480 \times 832$. Training is performed in two stages. In Stage 1, we optimize the forward flow map objective using AdamW~\cite{loshchilov2017decoupled} with a learning rate of $5 \times 10^{-5}$ and a global batch size of 32 for the 1.3B model and 16 for the 14B model, for 6,000 and 4,000 iterations, respectively. In Stage 2, we perform on-policy distillation by jointly optimizing the forward flow map and on-policy objectives with AdamW at a learning rate of $2 \times 10^{-6}$ for 800 iterations. In both stages, we use parameter-efficient fine-tuning with LoRA~\cite{hu2022lora} of rank 256.

\subsection{Evaluation Setting}
For text-to-video (T2V) evaluation, we adopt VBench~\cite{huang2023vbench}, which evaluates generation quality across 16 fine-grained dimensions grouped into two high-level categories: Quality score and Semantic score. For image-to-video (I2V) evaluation, we use VBench-I2V~\cite{huang2025vbench++} and report both the Quality score and the I2V score.

To ensure a fair comparison, we re-evaluate key counterparts under a unified protocol using the official VBench augmented prompts. These counterparts include the Wan2.1 base model~\cite{wan2025wan}, Self-Forcing~\cite{huang2025selfforcing}, and rCM~\cite{zheng2025large}, as well as community-trained 14B causal models including FastVideo~\cite{causalwan2025}, Krea-Realtime-14B~\cite{krea_realtime_14b}, and LightX2V~\cite{lightx2v}. Results for all other methods are taken directly from their original papers.
\subsection{Main Results}
\myPara{Quantitative Comparison.}
As shown in \cref{tab:t2v_comparison}, AnyFlow achieves strong few-step performance and continues to improve as the sampling budget increases. On the 14B bidirectional backbone, AnyFlow-Wan2.1-T2V-14B obtains 84.04 at 4 NFEs, outperforming rCM-Wan2.1-T2V-14B (83.73 at 4 NFEs). On the 14B causal backbone, AnyFlow-FAR-Wan2.1-14B reaches 84.05 at 4 NFEs and further improves to 84.41 at 32 NFEs, outperforming strong community counterparts.

For image-to-video results in \cref{tab:i2v_comparison}, AnyFlow-FAR-Wan2.1-14B achieves a VBench-I2V score of 87.87 using only 4 NFEs. This result is comparable to Wan2.1-I2V-14B with 50$\times$2 NFEs (87.71) and surpasses FastVideo-CausalWan2.2-A14B-Preview (86.82), showing that AnyFlow improves both visual quality and I2V faithfulness while retaining strong sampling efficiency.

\myPara{Qualitative Comparison.}
As shown in \cref{fig:14b_compare}, we provide qualitative comparisons at the 14B scale for both T2V and I2V generation. In causal T2V examples, FastVideo-Wan2.2-A14B-Preview exhibits blurry details, LightX2V-Wan2.1-14B-CausVid shows flickering artifacts, and Krea-Realtime-14B produces unrealistic motion in some clips. By comparison, AnyFlow-FAR-Wan2.1-14B delivers the best visual quality and motion with 4 NFEs. In bidirectional T2V examples, AnyFlow-Wan2.1-T2V-14B shows slightly better motion than rCM-Wan2.1-T2V-14B.

For I2V generation, AnyFlow reuses the causal generator through our non-uniform chunk partition and maintains good first-frame faithfulness with smooth motion transitions. Compared with Wan2.1-I2V-14B, AnyFlow-FAR-Wan2.1-14B shows comparable temporal stability and motion quality, while FastVideo-Wan2.2-A14B-Preview exhibits visible flickering and inconsistencies with the input image. Overall, these qualitative results are consistent with the quantitative trends and further support the effectiveness of AnyFlow in few-step sampling.

\begin{figure}[!tb]
    \centering
    \begin{subfigure}{0.45\textwidth}
        \centering
        \includegraphics[width=\textwidth]{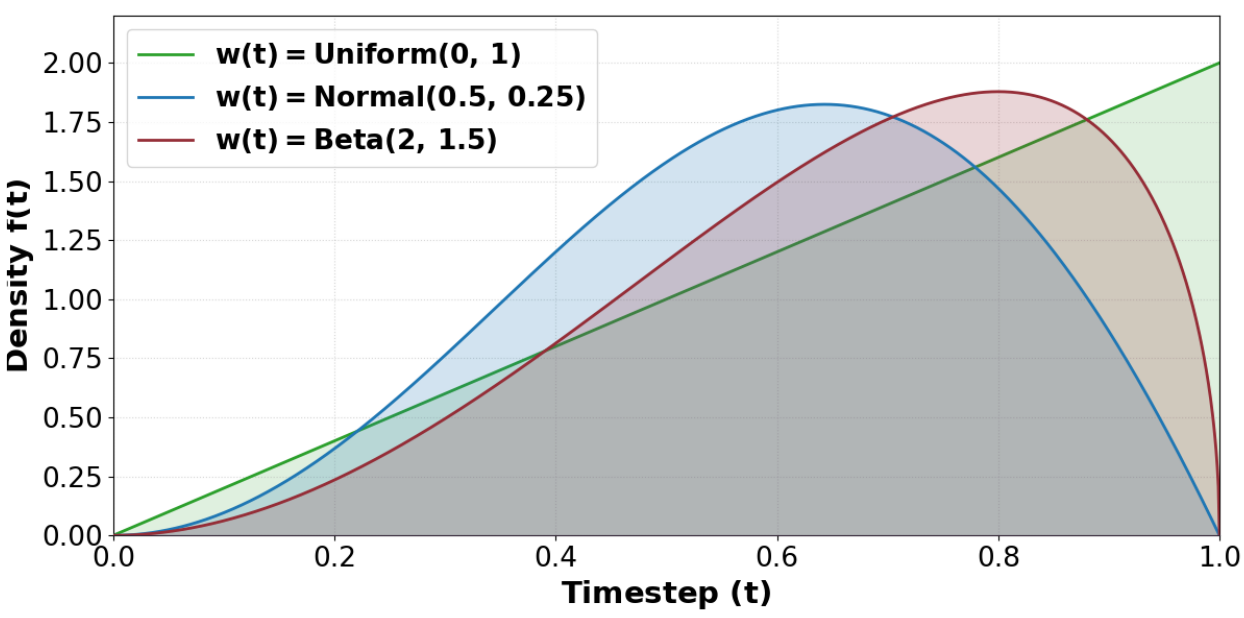}
        \caption{Timestep density $f(t)=p(t)\cdot w(t)$, where $p(t)=2t$.}
        \label{fig:vis_distribution}
    \end{subfigure}
    \hfill
    \begin{subtable}{0.53\textwidth}
        \centering
\renewcommand{\arraystretch}{1.55}
        \resizebox{\linewidth}{!}{\begin{tabular}{l|cc|cc}
    \toprule
    \multirow{2}{*}{$\mathbf{w(t)}$} & \multicolumn{2}{c|}{\textbf{Bidirectional}} & \multicolumn{2}{c}{\textbf{Causal}} \\
    \cmidrule(lr){2-3} \cmidrule(lr){4-5}
    & \textbf{4 NFEs} & \textbf{32 NFEs} & \textbf{4 NFEs} & \textbf{32 NFEs} \\
    \midrule
    $\text{Uniform}(0, 1)$    & 83.46 & 83.75 & \textbf{83.58} & 83.91 \\
    $\text{Normal}(0.5, 0.25)$  & 83.40 & 83.79 & 83.54 & 83.93 \\
    \rowcolor{mitblue}$\text{Beta}(2, 1.5)$   & \textbf{83.48} & \textbf{83.96} & 83.54 & \textbf{83.96} \\
    \bottomrule
\end{tabular}}
        \caption{Quantitative ablation (Vbench overall score) of the loss weight function $w(t)$.}
        \label{tab:ablation_time_sampler}
    \end{subtable}
    \caption{\textbf{Ablation Study of Loss Weight Function $w(t)$}. Among the different weighting strategies, $w(t) = \text{Beta}(2, 1.5)$ demonstrates the best performance across both architectures.}
\label{fig:ablation_time_sampler}
\end{figure}
\begin{figure}[!tb]
    \centering
    \begin{subfigure}{\textwidth}
        \centering
        \includegraphics[width=\textwidth]{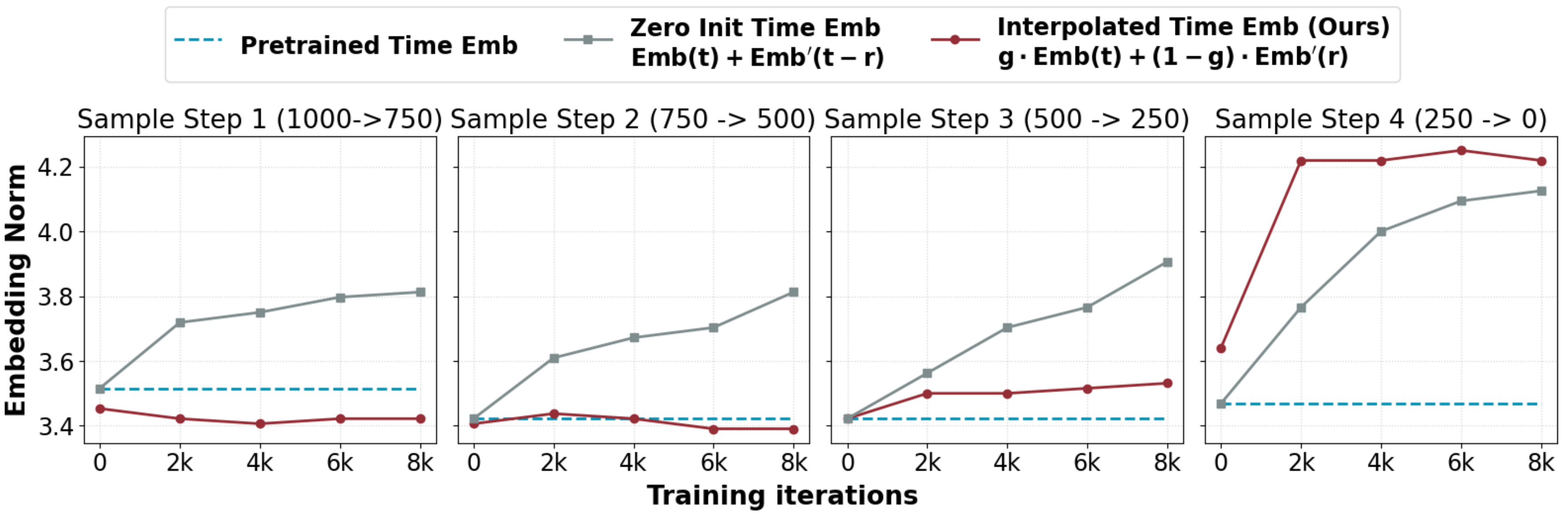}
        \caption{Comparison of embedding norms at different sample steps during training.}
        \label{fig:interpolated_time_norm}
    \end{subfigure}
    
    \begin{subfigure}{\textwidth}
        \centering
        \includegraphics[width=\textwidth]{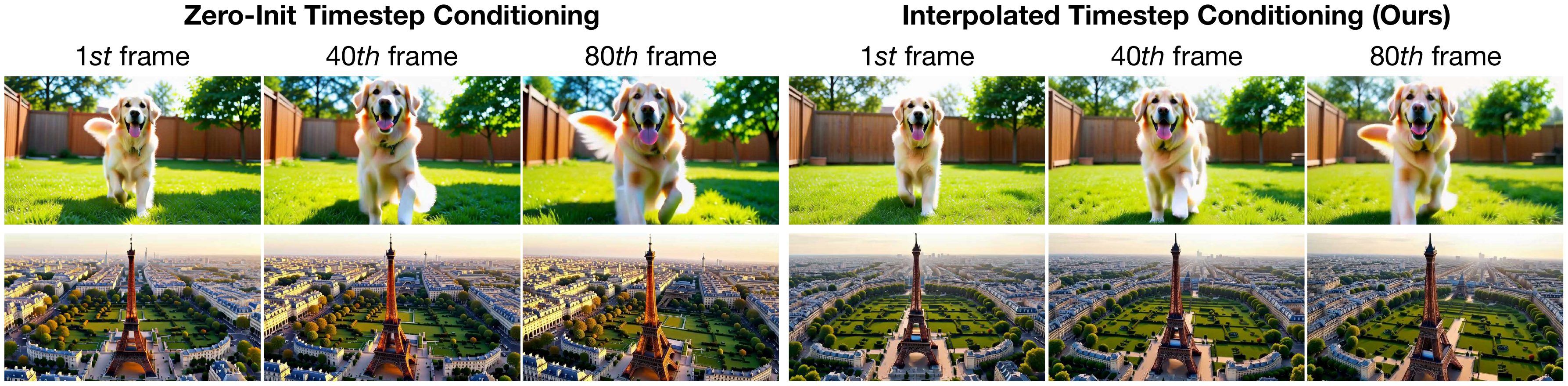}
        \caption{Visual comparison of zero-init timestep conditioning and interpolated timestep conditioning.}
        \label{fig:interpolated_time_vis}
    \end{subfigure}
    \caption{\textbf{Ablation Study of Timestep Conditioning}. Compared with zero-init timestep conditioning, the proposed interpolated method achieves embedding norms more consistent with pretrained embeddings, preventing over-saturated visual results.}
    \label{fig:ablation_interpolated_time}
\end{figure}

\subsection{Ablation Study}
\myPara{Time Sampler.}
We investigate different choices of the time sampler, with their induced densities over $t$ visualized in \cref{fig:vis_distribution}. The uniform weighting yields the worst test-time scalability, as reflected by its 32-step performance in \cref{tab:ablation_time_sampler}, likely because it mismatches the logit-normal time weighting used in pretraining. Among the other variants, assigning more probability mass to high-noise regions leads to better overall performance. Therefore, we fix $w(t)=\text{Beta}(2, 1.5)$ in all experiments.

\myPara{Interpolated Timestep Conditioning.} Incorporating an additional timestep embedding during post-training is non-trivial. A baseline design is to add a new embedding projection whose final layer is zero-initialized, as in TMD~\cite{nie2026transition}. However, under zero initialization, the model needs to amplify the influence of the $r$ embedding, which causes its norm to keep increasing during training and eventually leads to over-saturated results, as shown in \cref{fig:ablation_interpolated_time}. In contrast, our interpolated timestep conditioning preserves the boundary case $t=r$ for stability while avoiding an overly cold start when $t\neq r$. As a result, the embedding norm remains stable after roughly 2K training steps and stays much closer to that of the pretrained model, effectively suppressing the over-saturation effect.

\begin{figure*}[!tb]
  \centering
  \animategraphics[width=\linewidth]{8}{figures/videosrc/ablation_odeinit/}{00000}{00020}
  \vspace{-.3in}\caption{\textbf{Qualitative Ablation Study of ODE-Init and AnyFlow on Causal Video Generation.} Readers can \textcolor{magenta}{click and play} the video clips in this figure using Adobe Acrobat.}
  \label{fig:ablation_odeinit}
\end{figure*}

\myPara{Comparison to ODE-Init.}
We compare AnyFlow with Consistency ODE-Init for causal video generation in \cref{fig:ablation_odeinit}. For a fair comparison, we fine-tune the teacher on the same synthetic dataset, since we do not have access to the original training data of the Wan base model. After flow map training, AnyFlow produces clearer results than Consistency ODE-Init at both 4 and 32 NFEs, indicating that it better preserves the knowledge inherited from the flow-matching-pretrained base model. After on-policy distillation, AnyFlow further reduces test-time errors in few-step sampling and improves the overall sampling trajectory. We also observe that our method preserves the style and content of the pretrained model more faithfully.

\begin{table}[!tb]
    \centering
    \resizebox{0.85\linewidth}{!}{
    \begin{tabular}{l|ll}
        \toprule
        \multirow{2}{*}{\textbf{Method}} & \multicolumn{2}{c}{\textbf{Training Cost (s/iter on H100)}} \\
        \cmidrule{2-3}
         & \textbf{Causal} & \textbf{Bidirectional} \\
        \midrule
        \midrule
        \multicolumn{3}{c}{\textbf{Forward Training}} \\
        \midrule\midrule
        Standard Flow-Matching & 5.8 & 9.7 \\
        + Guidance-Fused Training & 7.8 & 12.3 \\
        \rowcolor{mitblue}+ Guidance-Fused Training + Differential Derivation Equation & 10.4 & 16.8 \\
        \midrule
        \midrule
        \multicolumn{3}{c}{\textbf{On-Policy Distillation}} \\
        \midrule\midrule
        Consistency Backward Simulation (4 Steps) & 45.9 & 41.8 \\
        \rowcolor{mitblue}Flow Map Backward Simulation (4 Steps) & 53.1 (\textcolor{red}{+15.7\%}) & 51.2 (\textcolor{red}{+22.5\%}) \\\midrule
        Consistency Backward Simulation (16 Steps) & 93.8 & 96.6 \\
        \rowcolor{mitblue}Flow Map Backward Simulation (16 Steps) & 53.1 (\textcolor{green!50!black}{-43.4\%}) & 51.2 (\textcolor{green!50!black}{-47.0\%}) \\
        \bottomrule
    \end{tabular}}
      \caption{\textbf{Training Cost Breakdown of AnyFlow.} Costs are measured on Wan2.1-1.3B-T2V with batch size 16 on a node with 8$\times$NVIDIA H100 GPUs. During on-policy distillation, we evaluate the upper-bound cost (comprising one generator and one discriminator update) for the  simulation trajectories.}

    \label{tab:train_cost}
\end{table}
\myPara{Training Cost Breakdown.} \cref{tab:train_cost} reports the training cost of each component in AnyFlow, measured in seconds per iteration on 8$\times$H100 GPUs. In the forward training stage, guidance-fused training and the differential derivation equation introduce additional overhead compared with standard flow matching, but the overall cost remains practical for large-scale training. In the on-policy distillation stage, our flow map backward simulation incurs a slightly higher cost than consistency backward simulation at 4 steps, specifically 15.7\% more for the causal model and 22.5\% more for the bidirectional model, because we backpropagate through the full rollout chain. However, our method becomes significantly more efficient when simulating larger step counts. At 16 steps, it reduces the training cost by 43.4\% for the causal model and 47.0\% for the bidirectional model compared with consistency backward simulation, thanks to the shortcut transitions learned by the flow map.

\section{Conclusion}
We present AnyFlow, the first any-step video diffusion distillation framework based on a two-time flow map formulation. By learning transitions between arbitrary time pairs, AnyFlow supports a wide range of sampling budgets within a single model. Built on large pretrained video diffusion backbones, it combines an improved forward flow map training recipe with flow map backward simulation for on-policy distillation, thereby reducing discretization error and exposure bias during sampling. Across both bidirectional and causal architectures, and across model scales from 1.3B to 14B, AnyFlow consistently matches or outperforms strong consistency-based counterparts in the few-step regime while continuing to improve as the sampling budget increases. These results highlight a practical and scalable path toward high-quality any-step video generation.

\myPara{Limitations.} The primary limitation of our method is its reliance on external datasets for flow map training. Even when synthetic data are used, the training distribution may still differ from that of the base model, which can introduce mild distribution shift, such as smoother textures. This issue could be mitigated by applying AnyFlow with the same data used to pretrain the base model.

\myPara{Future Work.} Although AnyFlow establishes the first any-step, flow-map-based video diffusion distillation framework, there is still substantial room for improving the training methodology. One important direction is to develop more effective and stable forward flow map training strategies to improve robustness across different NFE regimes. In addition, because AnyFlow provides a scalable recipe for learning any-step causal video diffusion from both data and a teacher model, a natural next step is to extend the framework to autoregressive long-video generation with dedicated long-video training.

{
  \small
  \bibliographystyle{unsrt}
  \bibliography{arxiv}
}

\clearpage

\end{document}